\documentclass{article}
\usepackage[preprint]{template}
\usepackage[utf8]{inputenc} 
\usepackage[T1]{fontenc}    
\usepackage{hyperref}       
\usepackage{url}            
\usepackage{booktabs}       
\usepackage{amsfonts}       
\usepackage{nicefrac}       
\usepackage{microtype}      
\usepackage{xcolor}         
\usepackage{graphicx}
\usepackage{amsmath}
\usepackage{amssymb,graphicx}

\title{Align2Act: Instruction-Tuned Models for Human-Aligned Autonomous Driving}

\author{
  Kanishkha Jaisankar \\
  Masters in Data Science \\
  New York University \\
  \texttt{kj2675@nyu.edu} \\
  \And
  Sunidhi Tandel \\
  Masters in Computer Engineering \\
  New York University \\
  \texttt{sdt9243@nyu.edu} \\
}

\begin{document}

\maketitle
\begin{abstract}
Motion planning in complex scenarios is a core challenge in autonomous driving. Conventional methods apply predefined rules or learn from driving data to generate future trajectories. Recent approaches leverage the reasoning capabilities of large language models (LLMs) for decision-making in driving scenarios. Despite promising results, it remains unclear whether LLMs truly capture the underlying human logic of driving. In this paper, we propose \textbf{Align2Act}, a motion planning framework that transforms instruction-tuned LLMs into interpretable planners aligned with human behavior. We derive driving instructions based on human reasoning patterns (e.g., anticipate hazards, yield near intersections) and traffic rules (e.g., stop at red lights, stay within lane boundaries). We then employ an \textbf{Align2ActChain} module to guide step-by-step reasoning, producing both the rationale and the trajectory. Align2ActDriver supports both instruction injection and learning from large-scale driving data, enhancing interpretability and behavioral alignment. Different from prior works focused on synthetic or open-loop settings, we evaluate our method on the real-world nuPlan closed-loop benchmark (Test 14-random), demonstrating improved planning quality and human-likeness.
Our code is publicly available at : \texttt{\textcolor{blue}{\href{https://github.com/Jkanishkha0305/Align2Act}{https://github.com/Jkanishkha0305/Align2Act}}}.
\end{abstract}

\section{Introduction}
Motion planning in autonomous driving is a core challenge in Embodied AI, requiring agents to perceive, reason, and act in dynamic environments \cite{li2024driving}. Unlike traditional software systems, autonomous vehicles must process complex multi-agent interactions, predict future states, and execute safe behaviors while adapting to unpredictable road conditions \cite{dongbevlm}. Traditional motion planning methods are typically categorized into rule-based approaches, which rely on predefined heuristics such as optimization-based trajectory planning or finite-state machines, and learning-based models, which leverage deep neural networks for decision-making \cite{sima2024drivelm, li2024driving}. While rule-based methods provide interpretability, they struggle with adaptability, whereas learning-based models exhibit flexibility but often lack robustness and causal reasoning, particularly in unseen scenarios. Both approaches face difficulties in handling rare edge cases, necessitating a more cognitively grounded alternative \cite{shao2024lmdrive}.  

Recent research suggests that Large Language Models (LLMs) could offer a novel perspective by incorporating knowledge synthesis, causal reasoning, and multi-step planning \cite{zhang2024instruct}. LLMs can integrate traffic rules, human-like heuristics, and adaptive reasoning into a single framework \cite{pengsystem}, but their direct application to motion planning remains uncertain. Key challenges include whether LLMs possess an embodied understanding beyond static reasoning \cite{dubey2024llama}, their ability to replicate human-like decision-making from structured driving data \cite{yang2024qwen2}, and ensuring that their predictions remain interpretable, safe, and aligned with real-world constraints \cite{thakkar2023comprehensive}. Since LLMs lack built-in priors for physical interactions, their decisions are prone to hallucinations in real-world settings, making structured alignment mechanisms necessary to enforce safety constraints while maintaining adaptability \cite{mo2024fine}. The integration of explicit reasoning steps within the motion planning pipeline is critical, where planning should be decomposed into sub-tasks such as collision avoidance, rule compliance, and situational awareness. This instruction-driven approach enhances interpretability and ensures safer, more transparent decision-making in complex environments \cite{bai2022training}.

\begin{figure}[h]
    \centering
    \includegraphics[width=0.8\linewidth]{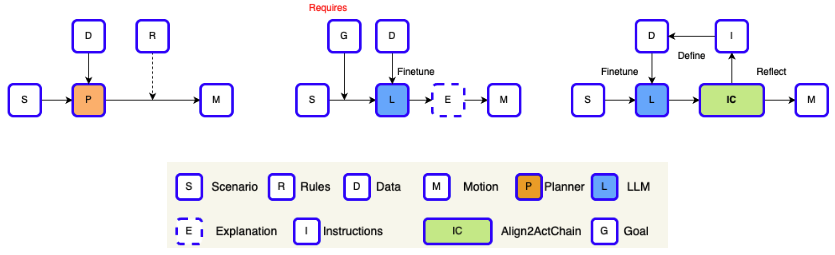}
    \caption{\footnotesize The Align2Act framework:
    \textbf{(D)ata} and \textbf{(S)cenario} inputs are processed through \textbf{(R)ules} and \textbf{(L)LM} reasoning to generate \textbf{(M)otion} plans via the \textbf{(IC) Align2ActChain}. 
    \textbf{(P)lanner} and \textbf{(E)xplanation} modules ensure interpretability, while \textbf{(I)nstructions} and \textbf{(G)oal} alignment refine human-like behavior.}
    \label{fig:framework}
\end{figure}

\section{Related Work}

We reviewed prior work in three areas to position our approach against existing benchmarks and gaps. 

\textbf{LLM-Based Planning Agents.}
LLM agents have demonstrated strong planning capabilities across simulated and physical domains. For example, Voyager was the first LLM-powered agent to autonomously explore and complete the Overworld tech tree in Minecraft, showcasing long-term planning and skill chaining \cite{zhu2023ghost}. LLMs have also enabled the simulation of a virtual town with 25 memory-augmented agents capable of reflection, planning and coordination \cite{park2023generative}. They have facilitated multi-agent collaboration through modular frameworks \cite{zhang2023building} and have been fine-tuned for semantically grounded robotic control in real-world settings \cite{brohan2023rt}. These advances make LLMs promising for high-level planning in autonomous driving.

\textbf{Motion Planning in Autonomous Driving without LLMs.}
Before LLMs, motion planners were primarily rule-based, learning-based, or hybrid. Rule-based models like IDM use heuristics for safe vehicle following \cite{treiber2000congested}. Learning-based approaches such as End-to-End driving \cite{bojarski2016end}, Conditional Imitation Learning \cite{codevilla2018end} and UrbanDriver \cite{scheel2022urban} learn driving strategies from real-world data. PlanTF leverages Transformers to generalize under long-tail scenarios \cite{cheng2023rethinking}. Hybrid methods like PDM integrate rule-based logic with learning-based models for improved robustness \cite{dauner2023parting}. However, these systems lack the reasoning flexibility and semantic richness that LLMs offer.

\textbf{LLM-Based Motion Planning in Autonomous Driving.}
Recent work integrates LLMs directly into the planning pipeline. LanguageMPC introduces cognitive reasoning pathways and converts high-level decisions into executable commands \cite{sha2023languagempc}. GPT-Driver formulates planning as a language modeling problem using linguistic tokens for trajectory prediction \cite{mao2023gpt}. Driving-with-LLMs fuses vectorized scene inputs with pretrained LLMs to improve contextual understanding \cite{chen2023driving}. LMDrive \cite{shao2024lmdrive} pioneers end-to-end closed-loop control with LLMs and LLM-ASSIST \cite{sharan2023llm} enhances closed-loop planners with language-based reasoning. These methods offer improved adaptability, interpretability and safety in complex driving settings.

\section{Proposed Approach}

\subsection{Motion Planning as Language Modeling}

We formulate motion planning as a conditional language generation problem. Given a structured textual input describing the vehicle's context and planning goals, a large language model is used to generate both the final driving trajectory and the corresponding reasoning steps. The goal is to produce human-aligned, interpretable plans that are safe and context-aware.

Let \(T\) denote the desired trajectory, and \(X_t\) be the planning context at time step \(t\). This input includes environmental observations \(O\), the ego vehicle's current state \(S_t\), motion system metadata \(S_y\), and planning instructions \(I\):

\begin{equation}
X_t = \{O, S_t, S_y, I\}
\end{equation}

This formulation enables the planner to reason over a combination of physical state and semantic instruction. Rather than predicting a trajectory directly from numerical features, we use language to encode the spatial-temporal context, system constraints, and behavioral intent.

The output \(Y_t\) consists of two components: the generated trajectory \(T\) and a structured reasoning trace, referred to as the \textbf{Align2ActChain}, denoted by \(I_c\). This intermediate chain articulates the model’s step-by-step logic in human-readable terms, providing interpretability and post-hoc traceability:

\begin{equation}
Y_t = \{I_c, T\}
\end{equation}

To generate this output, we pass the input \(X_t\) through a transformation function \(F\), which produces log-probabilities over the model vocabulary. These log-probabilities quantify the model's confidence in the next-token prediction during decoding:

\begin{equation}
Y_{\log} = F(X_t)
\end{equation}

This probabilistic output is then post-processed using a decoding mechanism that balances randomness and confidence. Specifically, temperature scaling \(T(\cdot, t)\) adjusts the sharpness of the distribution, followed by the Softmax function \(S\) to obtain probabilities, and finally top-\(p\) sampling \(P\) to select high-likelihood completions:

\begin{equation}
Y_t = P\left(S\left(T(Y_{\log}, t)\right), p\right)
\end{equation}

Here, the temperature parameter \(t\) modulates output diversity—higher values encourage exploration, while lower values enforce determinism. The top-\(p\) parameter defines a cumulative probability threshold for sampling, ensuring only the most plausible outputs are considered. This decoding process generates structured plans that balance safety, feasibility, and alignment with human-like decision-making.

\subsection{Instruction-Based Behavior Alignment}

To teach the model human-aligned behavior, we employ imitation learning via prompt-based supervision. The input \(X_t\) is decomposed into two parts: the instruction context \(I_t\) and the scene-based planning input \(I_p\):

\begin{equation}
X_t = \{I_t, I_p\}
\end{equation}

The instruction context \(I_t\) includes a description of the vehicle’s motion system \(S_y\) and the desired planning behavior \(I\):

\begin{equation}
I_t = \{S_y, I\}
\end{equation}

The motion system description \(S_y\) encodes vehicle-specific properties such as heading convention, yaw orientation, vehicle dimensions (modeled as quadrilaterals), and dynamic parameters like velocity and acceleration. The planning instructions \(I\) define task-level goals, such as "turn right" or "decelerate to yield," enabling the model to reason semantically rather than only kinematically.

The planning input \(I_p\) contains the ego-centric scene representation, composed of environment observations \(O\) and the ego vehicle’s state \(S_t\):

\begin{equation}
I_p = \{O, S_t\}
\end{equation}

Observations \(O\) include object categories (vehicles, cyclists, pedestrians), their relative positions, motion predictions, and semantic map elements such as lane lines, traffic lights, and speed limits. The self-state \(S_t\) includes historical trajectory information and the vehicle’s current kinematic state, which together help the model plan smooth transitions.

\subsection{Align2ActChain: Structured Reasoning for Interpretability}

The output of our model includes both the trajectory \(T\) and the Align2ActChain \(I_c\), which is a structured explanation of the decision-making process:

\begin{equation}
Y_t = \{I_c, T\}
\end{equation}

The \textbf{Align2ActChain} decomposes the planning process into four stages that reflect typical human reasoning in driving scenarios.

\paragraph{Preliminary Planning:} This step identifies a high-level maneuver such as continuing along the current lane, initiating a lane change, or preparing for a turn. It provides the initial semantic intent based on spatial context and scenario geometry.

\paragraph{Collision Prediction:} The planner forecasts the future positions of surrounding agents and evaluates their proximity to the ego vehicle. Agents within 3 meters are flagged as potential hazards, while those within 1.5 meters are considered critical, requiring immediate action to ensure safety.

\paragraph{Traffic Context Assessment:} This stage incorporates external constraints such as traffic light states, speed limits, and lane boundary conditions. Red signals prompt a full stop, yellow signals indicate caution, and green signals allow movement. Speed regulations are enforced by suppressing acceleration near or above the limit. Lane boundary violations are also considered during maneuver selection.

\paragraph{Final Action Integration: } The model synthesizes all prior reasoning to determine a safe and contextually appropriate driving action. This final decision is then mapped to a continuous trajectory \(T\), which is executed in simulation for downstream evaluation.

Trajectory \(T\) is represented as a sequence of poses over time, each with \(x\), \(y\), and \(\theta\) (yaw) coordinates:

\begin{equation}
T = \{(x_1, y_1, \theta_1), (x_2, y_2, \theta_2), \ldots, (x_n, y_n, \theta_n)\}
\end{equation}

This structured output is compatible with nuPlan’s simulation framework and can be executed via trajectory tracking controllers for open-loop and closed-loop evaluation.

\subsection{Align2ActDriver: Model Architecture and Fine-Tuning}

The Align2ActDriver framework uses LLaMA-2-7B as the backbone model and is fine-tuned using Low-Rank Adaptation (LoRA)~\cite{hu2022lora}, which enables efficient parameter updates with minimal memory overhead. The input sequence \(X_t\) is first passed through a preprocessing function \(P\), followed by an embedding layer \(E\) that transforms it into high-dimensional vector space:

\begin{equation}
X = E(P(X_t))
\end{equation}

This embedding \(X\) is then fed into the model \(F_m\), which includes attention modules and normalization. The output is a vocabulary-sized logit vector representing next-token probabilities:

\begin{equation}
Y_{\log} = F_m\left(X, \text{Att}(X, A, B, b), w R(X)\right)
\end{equation}

In this formulation, \(\text{Att}(X, A, B, b)\) refers to the attention mechanism, augmented by LoRA matrices \(A\) and \(B\), which act as low-rank approximations that reduce the number of trainable parameters. The bias term \(b\) and the RMSNorm function  with weight \(w\) help stabilize training and improve convergence. Only LoRA-specific parameters parameters (\(A, B, b, w\)) are updated during training, while the base LLaMA weights remain frozen, ensuring computational efficiency and performance.

This architecture enables scalable fine-tuning of large models for faster adaptation to the motion planning domain without requiring full retraining. The result is a planner that generates interpretable, human-aligned driving behavior based on high-level textual instructions and scene context.

\section{Experiments}

\subsection{Dataset and Preprocessing}
We conduct our experiments using the publicly available \textbf{nuPlan} dataset, which provides a large-scale, high-fidelity set of autonomous driving scenarios collected in diverse urban environments. The dataset comprises complex real-world driving data, including diverse traffic participants, signalized intersections, varying road geometries, and dynamic objects such as pedestrians and cyclists.

Each scenario in nuPlan is represented over a temporal window, containing ego vehicle states, surrounding agents, map information, and sensor-derived features. Following the methodology of PlanTF~\cite{jcheng2023plantf}, we utilized feature builders to extract the past 2 seconds of ego vehicle motion and predict the next 8 seconds, sampled at a temporal resolution of 0.1 seconds. However, to balance computational efficiency with predictive accuracy, our experiments used a coarser resolution of 0.5 seconds during training and inference.

To ensure coordinate invariance and facilitate generalization, all positional data are transformed such that the ego vehicle is positioned at the origin \((0, 0)\) of the coordinate frame with zero yaw. All surrounding agents' positions and yaw angles are converted relative to the ego-centric frame. A maximum of 32 surrounding agents are included per scenario, prioritized by spatial proximity.

In addition to numerical features, our approach incorporates structured language-based instructions. For instruction fine-tuning, we construct natural language prompts that describe the scenario context, intended maneuvers, and constraints. These instructions are encoded and concatenated with trajectory and state information to condition the LLM. The final dataset includes approximately one million labeled instruction-driven scenarios curated from the nuPlan Mini Split.

\subsection{Experiment Setup}
We fine-tuned the \textbf{LLaMA-2-7B} language model using \textbf{LoRA (Low-Rank Adaptation)}\cite{hu2022lora} to efficiently adapt it for interpretable motion planning. We selected \textbf{LLaMA-2-7B} over other comparable models (e.g., Mistral-7B, Falcon-7B) due to its open-source availability and seamless compatibility with \textbf{LoRA} fine-tuning on resource-constrained hardware such as the NVIDIA T4 GPU.  The base weights, tokenizer, and configuration files (including \texttt{params.json}) were sourced from \textbf{HuggingFace} after receiving approval from Meta. Training was conducted on a single NVIDIA T4 GPU with 8 CPU cores using the Lightning AI Studio platform. Training spanned approximately 12 hours over \textbf{6 epochs}, including a warm-up phase during the first epoch to stabilize training dynamics.

The LoRA configuration employed a rank of 16, with tuning applied to both the \textbf{bias} and \textbf{normalization layers}. This setup limited the number of trainable parameters to approximately \textbf{41.6 million}, significantly reducing the computational overhead compared to full fine-tuning. The model was trained using \texttt{bf16} precision to optimize GPU memory usage without sacrificing numerical stability.

We utilized the \texttt{llama-2-accessory} toolkit for training, which includes utilities for LoRA, tokenizer integration, and low-rank updates. The tokenizer was based on \textbf{SentencePiece}, with a vocabulary of 32,000 tokens and a maximum input sequence length of 512 tokens. For generation tasks, we allowed input prompts of up to 12,288 tokens to accommodate detailed instruction planning.

The learning rate was initialized at \(5 \times 10^{-5}\), decaying to a minimum value of \(5 \times 10^{-6}\), with a weight decay factor of 0.02 to prevent overfitting. We used gradient clipping at a value of 2 to stabilize training and accumulated gradients over 2 steps to simulate a larger batch size, as the batch size was constrained to 1 due to GPU limitations. 

During inference, generation was performed with a temperature of 0 for deterministic decoding and a top-\(p\) sampling threshold of 0.75 to balance exploration and certainty. Attention computations were accelerated using \textbf{FlashAttention} to reduce memory consumption, while \textbf{APEX} enabled mixed-precision training with improved throughput.

Our training pipeline used data parallelism via Scaled Dot Product Attention (SDP) and was optimized for low-memory, single-GPU execution. This configuration enabled scalable fine-tuning and inference, making the Align2ActDriver framework suitable for efficient motion planning experimentation.

\begin{figure}[h]
    \centering
    \includegraphics[width=0.6\linewidth]{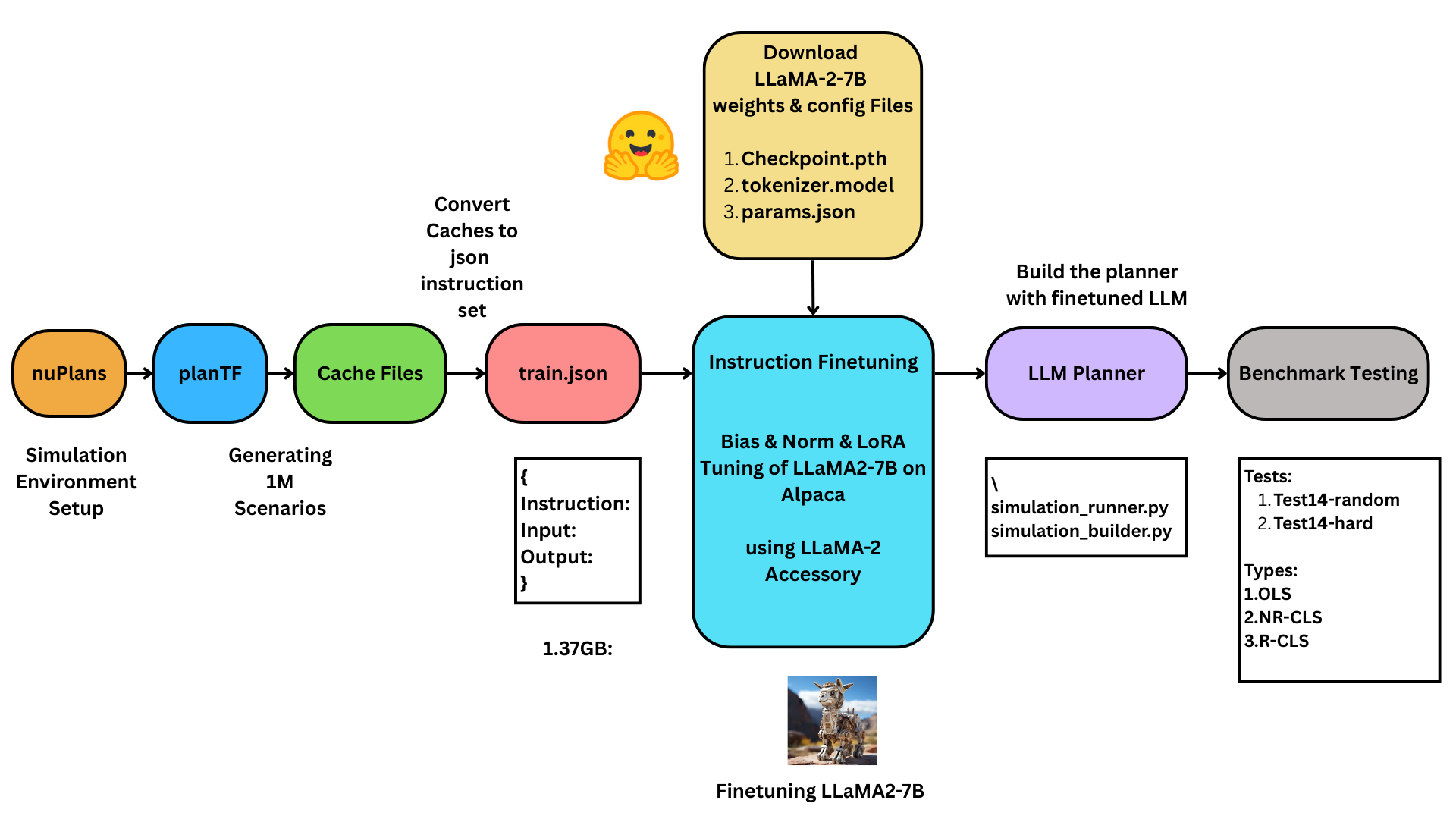}
    \caption{End-to-end project workflow: from data acquisition and simulation setup to training, fine-tuning, and benchmarking with LLM-based planners.}
    \label{fig:project_workflow}
\end{figure}

\subsection{Evaluation Metrics}

To evaluate the quality, safety, and human-likeness of the generated driving plans, we adopt the official evaluation protocol provided by the nuPlan benchmark. This framework includes both open-loop and closed-loop metrics, designed to assess spatial accuracy, trajectory feasibility, rule compliance, and interaction with dynamic agents.

\textbf{Open-Loop Score (OLS)} measures the planner's ability to imitate human trajectories without simulation feedback. It aggregates multiple sub-metrics, including the \textit{Average Distance Error (ADE)}, which computes the mean Euclidean distance between predicted and ground-truth trajectories over the planning horizon, and the \textit{Final Distance Error (FDE)}, which captures deviation at the final predicted timestep. Additionally, orientation accuracy is evaluated using \textit{Average Heading Error (AHE)} and \textit{Final Heading Error (FHE)}. The \textit{Missing Rate} accounts for the proportion of trajectories with errors exceeding a predefined threshold, penalizing plans with poor spatial alignment.

\textbf{Nonreactive Closed-Loop Score (NR-CLS)} evaluates trajectory execution within a simulated environment, where the ego vehicle operates in a static world without reactive background agents. This score reflects the planner's ability to generate smooth, rule-compliant trajectories while maintaining lane discipline and avoiding illegal maneuvers.

\textbf{Reactive Closed-Loop Score (R-CLS)} extends the NR-CLS evaluation by incorporating simulated traffic participants governed by the Intelligent Driver Model (IDM). This setting introduces interactive complexity, requiring the planner to respond appropriately to leading or crossing vehicles, pedestrians, and other dynamic actors. R-CLS therefore reflects both motion feasibility and interactive robustness.

All three scores range from 0 to 100, with higher values indicating better planning performance. Together, they offer a comprehensive assessment of trajectory prediction quality, interpretability, and real-world deployability.

\subsection{Main Results}
We evaluated our method, \textbf{Align2Act}, on two benchmark splits from the nuPlan dataset: \textit{Test14-random} and \textit{Test14-hard}. These benchmarks include diverse and challenging scenarios representative of both common urban driving tasks and edge cases. Our planner was deployed within the nuPlan simulation environment, using an LQR (Linear Quadratic Regulator) controller for trajectory execution, and a kinematic vehicle model to update state transitions.

As shown in Table~\ref{tab:main_results}, Align2Act achieved strong results across all metrics. On \textit{Test14-random}, we obtained an \textbf{OLS of 85.17}, with NR-CLS and R-CLS scores of \textbf{70.31} and \textbf{66.96}, respectively. On the more challenging \textit{Test14-hard} set, Align2Act achieved an OLS of \textbf{81.12}, demonstrating the model’s generalization to difficult edge cases. Compared to rule-based and hybrid planners such as IDM and PDM, our approach consistently outperformed in open-loop settings and remained competitive in closed-loop evaluations.

\begin{table}[h]
\centering
\caption{Comparison of Align2Act variants with baseline planners on the nuPlan benchmark. Metrics are OLS, NR-CLS, and R-CLS on Test14-random and Test14-hard splits.}
\label{tab:main_results}
\begin{tabular}{l|ccc|ccc}
\toprule
\textbf{Planner} & \multicolumn{3}{c|}{\textbf{Test14-random}} & \multicolumn{3}{c}{\textbf{Test14-hard}} \\
 & OLS & NR-CLS & R-CLS & OLS & NR-CLS & R-CLS \\
\midrule
\textbf{Expert (LogReplay)} & \textbf{100.0} & \textbf{94.03} & \textbf{75.86} & \textbf{100.0} & \textbf{85.96} & \textbf{68.80} \\
\midrule
\textbf{Rule-based} \\
IDM~\cite{li2024ego} & 34.15 & 70.39 & 72.42 & 20.07 & 56.16 & 62.26 \\
PDM-Closed~\cite{dauner2023parting} & 46.32 & 90.05 & 91.64 & 26.43 & 65.07 & 75.18 \\
\midrule
\textbf{Hybrid} \\
GameFormer~\cite{huang2023gameformer} & 79.35 & 80.80 & 79.31 & 75.27 & 66.59 & 68.83 \\
PDM-Hybrid~\cite{dauner2023parting} & 82.21 & \textbf{90.20} & \textbf{91.56} & 73.81 & 65.95 & \textbf{75.79} \\
\midrule
\textbf{Learning-based} \\
PlanCNN~\cite{caesar2021nuplan} & 62.93 & 69.66 & 67.54 & 52.40 & 49.47 & 52.16 \\
UrbanDriver~\cite{scheel2022urban}& 82.44 & 63.27 & 61.02 & 76.90 & 51.54 & 49.07 \\
GC-PGP~\cite{hallgarten2023prediction} & 77.33 & 55.99 & 51.39 & 73.78 & 43.22 & 39.63 \\
PDM-Open~\cite{dauner2023parting} & \textbf{84.14} & 52.80 & 57.23 & \textbf{79.06} & 33.51 & 35.83 \\
PlanTF~\cite{jcheng2023plantf} & \underline{87.07} & \underline{86.48} & \underline{80.59} & \underline{83.32} & \underline{72.68} & \underline{61.70} \\
\midrule
\textbf{LLM-based (Ours)} \\
Align2Act (LLaMA2 without chain) & 61.23 & 46.91 & 42.47 & 68.02 & 55.14 & 50.31 \\
Align2Act (InstructionChain) & \underline{85.17} & 70.31 & 66.96 & \underline{81.12} & 57.37 & 52.95 \\
\bottomrule
\end{tabular}
\end{table}

\begin{figure}[h]
    \centering
    \includegraphics[width=0.7\linewidth]{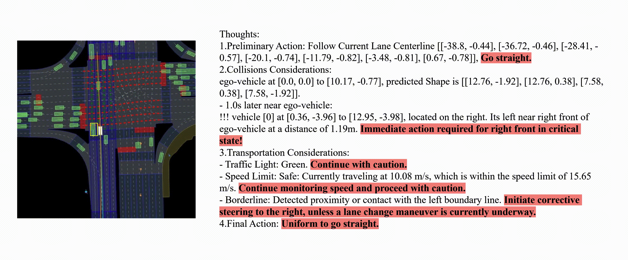}
    \caption{\footnotesize Visualization of Align2Act's reasoning process for a lane-following scenario. 
    The model's decision-making chain is shown: 
    (1) Preliminary action selection (blue trajectory), 
    (2) Collision risk assessment (red bounding boxes for critical agents), 
    (3) Traffic rule compliance (green light status and speed monitoring), and 
    (4) Final action execution (bold trajectory with corrective steering). 
    The ego vehicle (yellow) maintains safe distances while adhering to lane boundaries.}
    \label{fig:reasoning_vis}
\end{figure}

\subsection{Ablation and Analysis}

We conducted a series of ablation studies to evaluate the impact of reasoning structure, scenario diversity, temporal context, and input representations on planning performance. Table~\ref{tab:ablation-results}.

\textbf{Structured Reasoning.} To measure the effect of reasoning, we removed the Align2ActChain module and evaluated a flat-input baseline. Across all metrics, performance declined. For example, in a 5-scenario test configuration, the Open-Loop Score (OLS) dropped from 67.69 (with reasoning) to 59.91 (without), and the Reactive Closed-Loop Score (R-CLS) fell from 41.25 to 37.35. These results highlight the importance of structured reasoning for accurate and interpretable planning.

\textbf{Scenario Diversity.} We further tested generalization by expanding the number of training scenarios from 5 to 7. This increased diversity improved performance even without the reasoning chain (OLS: 61.23), and led to stronger gains when combined with Align2ActChain (OLS: 69.10, R-CLS: 43.80), confirming that broader scenario exposure strengthens contextual understanding.

\textbf{Temporal Correlation vs. Diversity.} To assess the importance of temporality, we compared two models: one trained on 300,000 independent, non-temporal scenarios, and another trained on 20,000 sequences with 16 temporally correlated frames. The non-temporal model outperformed its temporal counterpart (OLS: 63.11 vs. 60.35, R-CLS: 24.83 vs. 22.10), suggesting that scenario diversity provides a stronger training signal than frame-level continuity under limited compute. These findings indicate that, when training budgets are constrained, maximizing environmental variety may offer better generalization than increasing temporal depth.

\textbf{Previous Trajectory Input.} We also tested using the planned trajectory from the previous frame as an additional input. This led to performance degradation across all metrics. We hypothesize that the model overfit to shallow mappings between consecutive outputs, neglecting environmental awareness and state reasoning.

\begin{table}[h]
\centering
\caption{Ablation results showing the impact of reasoning, temporality, and scenario scale. Best scores in bold.}
\label{tab:ablation-results}
\begin{tabular}{lcccc}
\toprule
\textbf{Model Variant} & \textbf{OLS} & \textbf{NR-CLS} & \textbf{R-CLS} \\
\midrule
No Reasoning Chain (5 scenarios)             & 59.91 & 38.45 & 37.35 \\
No Reasoning Chain (7 scenarios)              & 61.23 & 39.18 & 38.27 \\
With Align2ActChain (5 scenarios)             & 67.69 & 40.84 & 41.25 \\
With Align2ActChain (7 scenarios)             & 69.10 & 42.90 & 43.80 \\
Align2ActChain + 16 Temporal Frames + 20,000 scenes    & 60.35 & 22.90 & 22.10 \\
Prev. Trajectory Input (PT, no temp.) + 300,000 scenes   & 63.11 & 25.52 & 24.83 \\
\bottomrule
\end{tabular}
\end{table}

\section{Limitations}

While Align2ActDriver improves interpretability through structured reasoning, it underperforms conventional learning-based and hybrid planners in closed-loop settings. The Align2ActChain enhances transparency but lacks the robustness needed for real-time deployment.

Despite using LoRA for efficient fine-tuning, reliance on large language models leads to high inference latency and memory demands, limiting feasibility for on-vehicle applications without further optimization.

The manually crafted instruction format also presents scalability challenges, particularly in unseen or highly diverse scenarios. Generalization may require adaptive prompting or automated instruction generation.

Evaluation was limited to a subset of nuPlan. Broader benchmarks such as val14 could reveal generalization limits. Moreover, the absence of visual or LiDAR inputs restricts reasoning in perceptually complex environments.

\section{Conclusion}

We introduced Align2ActDriver, a motion planning framework that uses instruction-tuned language models to produce interpretable, human-aligned trajectories. By modeling planning as language generation, it enables structured reasoning through the Align2ActChain.

Evaluations on nuPlan show competitive open-loop results and coherent, rule-following behavior. Though closed-loop performance lags behind traditional planners, Align2ActDriver offers strong interpretability and alignment with human driving logic.

Ablation studies highlight the benefits of temporal context, scenario diversity, and reasoning chains. Visualizations confirm the model's capacity to respond to realistic driving cues.

Future work includes integrating visual inputs, reducing latency, and scaling to multi-agent and broader benchmarks like val14. This approach marks a step toward bridging generative reasoning with real-world autonomous control.


\bibliographystyle{unsrt}
\bibliography{template}

\end{document}